\documentclass[10pt,letterpaper]{article}

\usepackage{cogsci}

\cogscifinalcopy 

\usepackage{cmap}
\usepackage[T1]{fontenc}
\usepackage[american]{babel}
\usepackage{csquotes}
\usepackage{newtxtext,newtxmath}  

\usepackage[
  backend=biber,
  style=apa,
  natbib=true,
  eprint=true,
  annotation=false,
]{biblatex}
\usepackage{float} 

\usepackage{graphicx}
\usepackage{fontawesome5}
\usepackage{contour}
\usepackage[normalem]{ulem}

\contourlength{0.8pt}

\newcommand{\myuline}[1]{%
  \uline{\phantom{#1}}%
  \llap{\contour{white}{#1}}%
}

\usepackage[hidelinks]{hyperref}

\title{On periodic distributed representations using Fourier embeddings}

\author[1,2]{\mbox{Jakeb Chouinard (jakeb.chouinard@uwaterloo.ca)}}
\affil[1]{Department of Systems Design Engineering, University of Waterloo}
\affil[2]{Centre for Theoretical Neuroscience, University of Waterloo}

\begin{document}

\maketitle

\begin{abstract}
Periodic signals are critical for representing physical and perceptual phenomena. Scalar, real angular measures, e.g., radians and degrees, result in difficulty processing and distinguishing nearby angles, especially when their absolute difference exceeds $\pi$. We can avoid this problem by using real-valued, periodic embeddings in high-dimensional space. These representations also allow us to control the nature of their dot product similarities, allowing us to construct a variety of different kernel shapes. In this work, we aim of highlight how these representations can be constructed and focus on the formalization of Dirichlet and periodic Gaussian kernels using the neurally-plausible representation scheme of Spatial Semantic Pointers\footnote{Code implementing the discussed SSP representations is available in the \faGithub\,\,\myuline{\href{https://github.com/ctn-waterloo/sspspace/tree/main}{``sspspace'' Python package}} as kernel configurations of the ``CyclicSSPSpace'' encoder.}.

\textbf{Keywords:}
Distributed Representations; Fourier Features; Spatial Semantic Pointers; Periodic Embeddings
\end{abstract}

\section{Introduction}

The natural world has no lack of periodic signals and repeating phenomena, such as orientation angles, circadian rhythms, and musical patterns. Critically, most approaches to distributed embeddings, such as Random Fourier Features (RFFs; \cite{rahimi_random_2007}), focus on continuous feature representations, thereby mapping some infinite number line to some high-dimensional trajectory. Previous work has done little to characterize or analyse the periodic versions of these embeddings as well as their induced kernels (see \cite{voelker_short_2020, frady_computing_2021}). Compared to representing angles as scalar values, such as degrees or radians, a distributed embedding can avoid the sharp discontinuity than comes when comparing points near the limits of the repeating signal's period.

In this work, we focus on real-valued embeddings of the form $\phi(\boldsymbol{x})=\mathcal{F}^{-1}\left\{e^{jA\boldsymbol{x}}\right\}$ where $A\in\mathbb{R}^{d\times n}$ is conjugate symmetric, $\boldsymbol{x}\in\mathbb{R}^{n}$, and $j$ is the imaginary unit. These embeddings are known as Spatial Semantic Pointers \citep{komer_neural_2019}. We focus on these representations and their approach to embeddings due to their neural feasibility. Additionally, their compositional structure allows for a natural extension of these periodic representations to cognitive models.

While most $A$ sufficiently large matrices will not see repetition in their resultant kernels; it can be desirable to control the point at which a repetition would occur. As an example, head direction relative to one's environment or body direction can be thought of as an angular measure---a signal with a period of $2\pi$. A periodic embedding and its corresponding periodic kernel can be constructed by ensuring that all entries of $A$ have some common multiple, $t_{\circ}\in\mathbb{R}^{+}$, such that $\mathrm{mod}(kt_{\circ}A,2\pi)=\boldsymbol{0}\,\,\forall\,\,k\in\mathbb{Z}$. Notably, the ``distance'' between two orientation states, $\epsilon$ and $2\pi-\epsilon$ where $\epsilon$ is some small number, is poorly represented by their difference. That is to say, the actual distance between $\epsilon$ and $2\pi-\epsilon$ is much smaller than $|2\pi-2\epsilon|$. By contrast, distributed representations capture their adjacency using a dot product measure such that $\lim_{\epsilon\rightarrow0}\phi(\epsilon)\boldsymbol{\cdot}\phi(2\pi-\epsilon)=1$ and $\lim_{\epsilon\rightarrow0}\|\phi(\epsilon)-\phi(2\pi-\epsilon)\|_{2}=0$.

\section{Methods}

To construct some phase matrix, $A\in\mathbb{R}^{d\times n}$, with a feature-space period of $t_{\circ}\in\mathbb{R}^{+}$, all entries of $A$ must be integer multiples of period-scaled Fourier roots, $2\pi t_{\circ}^{-1}$. The set of possible entries for $A$, $\mathcal{S}_{\circ}$, can be written as:
\begin{equation}
    \mathcal{S}_{\circ}(t_{\circ})=\left\{2\pi n t_{\circ}^{-1}|n\in\mathbb{Z}\right\}
\end{equation}
with a corresponding probability mass function (PMF) $p[\omega]$. Since the PMF is a positive finite Borel measure in Fourier space, Bochner's theorem  demonstrates that there exists a positive definite function that defines a shift-invariant kernel over the feature space \citep{bochner_lectures_1959}. 

Let $A$ be a phase matrix realizing a periodic embedding with period $t_{\circ}$ whose column entries are i.i.d. uniformly sampled up to some frequency band $2\pi n t_{\circ}^{-1}B$:
\begin{equation}
    A\sim\Omega=\left\{2\pi n t_{\circ}^{-1}|n\sim\mathcal{U}[-B,B]\right\}
\end{equation}
where $B\in\mathbb{Z}^{+}$ and $\mathcal{U}[a,b]$ is uniform sampling of integers between $a$ and $b$ inclusive.

Let $\phi_{A}(\boldsymbol{x})$ and $\phi_{A}(\boldsymbol{y})$ be SSP embeddings of two distinct points. The inner product of these two vectors, $\phi_{A}(\boldsymbol{x})^{\top}\phi_{A}(\boldsymbol{y})$, approximates the similarity kernel between two points in $\mathbb{R}^{n}$: $K(\boldsymbol{x},\boldsymbol{y})$. Since the kernel is shift-invariant, $K(\cdot)$ can be written as function of the vector from $\boldsymbol{x}$ to $\boldsymbol{y}$ (or vice versa): $K(\boldsymbol{x}-\boldsymbol{y})$. Since the columns of A are i.i.d. and assumed to be from the same distribution, the $\mathbb{R}^{d}$ embeddings of individual features in $\mathbb{R}^{n}$ are independent, and their kernels multiply \citep{voelker_short_2020}:
\begin{equation}
    K(\boldsymbol{x})=\prod_{i=1}^{n}k(x_{i})
\end{equation}

An expression for $k(\cdot)$ can be found by taking the inverse Fourier transformation of the probability distribution function (PDF) from which entries of $A$ were sampled:
\begin{align}
    k(x)=&\mathcal{F}^{-1}\left\{p(\omega)\right\} \notag\\
    =&\int_{\Omega}e^{-j\omega x}p(\omega)d\omega
\end{align}

However, since $\Omega$ is a discrete set for cyclic embeddings, $p(\omega)$ transitions from a PDF to a PMF, $p[\omega]$, over the entries of $\Omega$. Notably, $A$ is conjugate symmetric to ensure $\phi\in\mathbb{R}^{d}$, resulting in every phase being matched with its conjugate:
\begin{align}
    k(x)=&\sum_{\omega\in\Omega}\left(e^{-j\omega x}+e^{j\omega x}\right)p[\omega] \notag\\
    =&\sum_{\omega\in\Omega}\cos(\omega x)p[\omega]
\end{align}
and for $A$ as defined in Equation 2:
\begin{align}
    k(x)=&\sum_{n=-B}^{B}\left(\cos(2\pi nt_{\circ}^{-1}x)\frac{1}{2B+1}\right) \notag\\
    =&\frac{1}{2B+1}\sum_{n=-B}^{B}\cos(2\pi nt_{\circ}^{-1}x)
\end{align}

This is a normalized Dirichlet kernel with period $t_{\circ}$, equivalently written as $\hat{D}_{B}(2\pi t_{\circ}^{-1}x)$. This is identical to a repeating sinc kernel---the non-repeating variation of which is the kernel induced by uniform sampling over a continuous distribution of phases. Returning to $K(\boldsymbol{x})$:
\begin{equation}
    K(\boldsymbol{x})=\prod_{i=1}^{n}\hat{D}_{B}(2\pi t_{\circ}^{-1}x_{i})
\end{equation}

Unfortunately, this $n$-dimensional kernel is not symmetric. The behaviour of the kernel for $\boldsymbol{x}=c\left[\frac{1}{\sqrt{n}},\ldots,\frac{1}{\sqrt{n}}\right]^{\top}$ is different from that of the kernel for $\boldsymbol{x}=c\left[1,0,\ldots,0\right]^{\top}$ where $c\in\mathbb{R}$. In the first case, the Dirichlet approximation is $\mathrm{sinc}^{n}(c)$, whereas in the second case, it is $\mathrm{sinc}(c)$. The product of sinc functions is highly oscillatory nature, as an individual sinc function changes signs several times near $0$ and has pronounced, negative lobes. These attributes make it less ideal for representing continuous, multidimensional feature spaces where equidistant features should be equally similar.

Given Equation 5 as a generalized form, we can consider alternate repeating kernel shapes as well. As an example, we consider Normal-like sampling of phases similarly constrained to some frequency band $B$:
\begin{equation}
    A\sim\Omega=\left\{2\pi nt_{\circ}^{-1}|n\sim\mathcal{N}\left[0,\sigma^{2},B\right]\right\}
\end{equation}
where $\sigma\in\mathbb{R}^{+}$ and $\mathcal{N}[0,a^{2},b]$ is a centered, pseudo-normal distribution over the integers from $-b$ to $b$ inclusive with shape parameter $a^{2}$---the construction of which is explained as follows. 

We calculate the PMF of $\mathcal{N}\left[0,\sigma^{2},B\right]$, $p[\omega]$, by evaluating a centred normal distribution's PDF with variance $\sigma^{2}$, $f_{\sigma^{2}}(x)$, for the possible entries of $A$ and normalizing:
\begin{align}
    p[\omega]=\frac{f_{\sigma^{2}}(\omega)}{\sum_{\omega\in \Omega}f_{\sigma^{2}}(\omega)}
\end{align}
resulting in the kernel:
\begin{align}
    k(x)=\frac{\sum_{\omega\in\Omega}\cos\left(\omega x\right)f_{\sigma^{2}}(\omega)}{\sum_{\omega\in\Omega}f_{\sigma^{2}}(\omega)}
\end{align}
or equivalently:
\begin{align}
    k(x)=\frac{\sum_{n=-B}^{B}\cos\left(2\pi nt_{\circ}^{-1}x\right)f_{\sigma^{2}}\left(2\pi nt_{\circ}^{-1}\right)}{\sum_{n=-B}^{B}f_{\sigma^{2}}\left(2\pi nt_{\circ}^{-1}\right)}
\end{align}

This is a periodic Gaussian similarity kernel. Supposing a sufficiently large $B$, the numerator and denominator of Equation 11 approximate a ratio of theta functions where each function is defined as:
\begin{equation}
    \theta(z,\tau)=\sum_{i=-\infty}^{\infty}\tau^{i^{2}}e^{2jiz}
\end{equation}
such that:
\begin{equation}
    k(x)=\frac{\theta\left(\pi t_{\circ}^{-1}x,g_{\sigma^{2}}\left(2\pi t_{\circ}^{-1}\right)\right)}{\theta\left(0,g_{\sigma^{2}}\left(2\pi t_{\circ}^{-1}\right)\right)}
\end{equation}
where $g_{\sigma^{2}}(x)=e^{-\frac{x^{2}}{2\sigma}}$. Considering $K(\boldsymbol{x})$ for a periodic Gaussian kernel in $n$-dimensions:
\begin{equation}
    \begin{aligned}
        K(\boldsymbol{x})=\frac{\prod_{i=1}^{n}\theta\left(\pi t_{\circ}^{-1} x_{i},g_{\sigma^{2}}\left(2\pi t_{\circ}^{-1}\right)\right)}{\theta\left(0,g_{\sigma^{2}}\left(2\pi t_{\circ}^{-1}\right)\right)^{n}}
    \end{aligned}
\end{equation}

\begin{figure}[h!]
    \centering
    \includegraphics[width=1\linewidth]{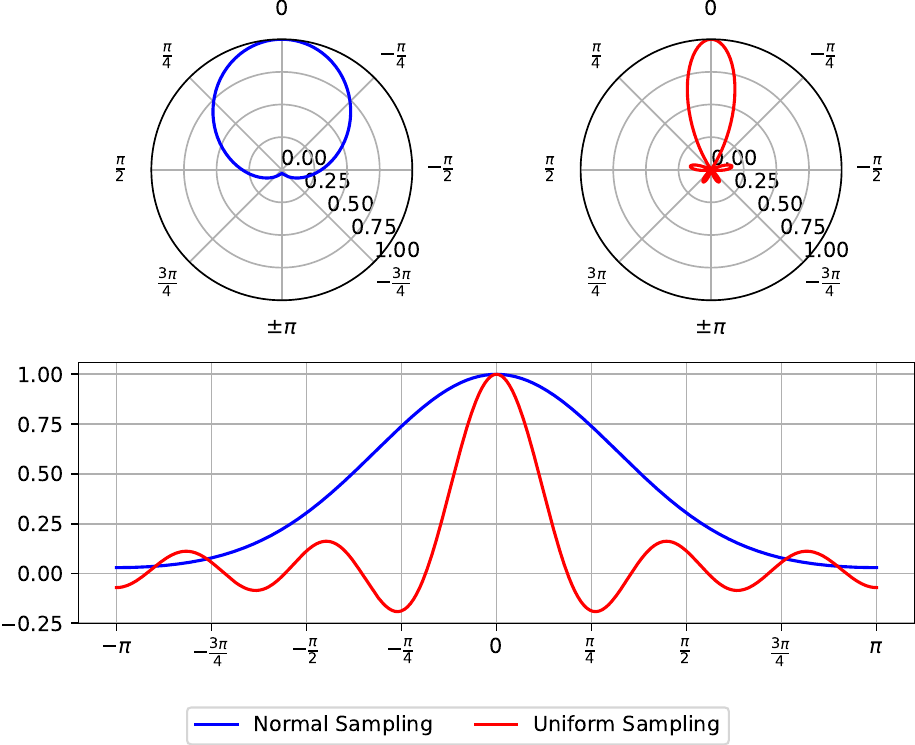}
    \caption{Average approximations of the normalized periodic Gaussian and normalized Dirichlet kernels using normal and uniform sampling for $A\in\mathbb{R}^{d\times1}$ respectively ($N=500$, $d=100$, $B=5$, $t_{\circ}=2\pi$, and $\sigma=1$).}
    \label{fig:r-theta-kernels}
\end{figure}

Equation 14 defines a product-of-Gaussians kernel where $K(\boldsymbol{x})$ produces a lattice of Gaussian-like similarity curves. While this multidimensional Gaussian kernel is radially asymmetric, similar to its Dirichlet equivalent, it is non-negative everywhere and smoother than the product-of-Dirichlets kernel. Figure \ref{fig:r-theta-kernels} provides a visualization of both the normalized Dirichlet kernel and the normalized repeating Gaussian kernel using uniform and normal sampling respectively.

\section{Conclusion}
In this brief account, we have laid out a way by which high-dimensional representations can be constructed such that they are cyclic. Further, we have characterized the repeating kernels for high-dimensional feature spaces and provided a novel insight into how one can sample different distributions to create varying kernel shapes. We hope to apply these techniques to cognitive models in the near future, using representations of direction, orientation, and even stimulus perception like colour to create models capable of exhibiting human-like behaviours.

\section{Acknowledgements}
The author would like to acknowledge and thank Nicole Sandra-Yaffa Dumont and Terrence Stewart for their technical assitance. The author would also like to thank Chris Eliasmith, Michael Furlong, Anna Penzkofer, and Madeleine Bartlett---all of whom were involved in discussions that motivated further examination of periodic distributed representations.

\printbibliography

\end{document}